\documentclass{article}

\usepackage{arxiv}

\usepackage[utf8]{inputenc} 
\usepackage[T1]{fontenc}   
\usepackage{hyperref}    
\usepackage{url}           
\usepackage{booktabs}      
\usepackage{amsfonts}       
\usepackage{nicefrac}    
\usepackage{microtype}      
\usepackage{lipsum}
\usepackage{changepage}
\usepackage{graphicx}
\usepackage{amsmath}
\usepackage{amsfonts}
\usepackage{amssymb}
\usepackage{makeidx}
\usepackage{algorithm,algorithmic}
\usepackage[switch]{lineno}
\usepackage{xcolor}
\usepackage{bm}
\usepackage[justification=centering]{caption}
\usepackage{amsfonts}
\usepackage[bbgreekl]{mathbbol}
\DeclareSymbolFontAlphabet{\mathbbm}{bbold}
\DeclareSymbolFontAlphabet{\mathbb}{AMSb}%
\usepackage{amsthm}
\theoremstyle{definition}
\newtheorem{theorem}{Theorem}

\title{A Survey on Universal Approximation Theorems}

\author{
  Midhun T. Augustine\\
  Automation Lab\\
  Indian Institute of Technology Bombay, India \\
  \texttt{midhunta30@gmail.com} \\
}
\begin{document}
\maketitle

\begin{abstract}
This paper discusses various theorems on the approximation capabilities of neural networks (NNs), which are known as universal approximation theorems (UATs). The paper gives a systematic overview of UATs starting from the preliminary results on function approximation, such as Taylor's theorem, Fourier's theorem, Weierstrass approximation theorem, Kolmogorov–Arnold representation theorem, etc. Theoretical and numerical aspects of UATs are covered from both arbitrary width and depth.

\end{abstract}


\section{Introduction}
 A \textit{neural network} (NN) or artificial neural network (ANN) is a network of artificial neurons arranged in layers \cite{bMP43,bSH}.
 The artificial neurons (also called perceptrons) are inspired by biological neurons in biological neural networks (BNNs)\cite{bEK}.  
 Biological neurons are the signal-processing units of BNN in the brain, similarly,  artificial neurons are data-processing units in  ANN. The rest of the paper discusses only ANN and artificial neurons which will be referred to simply by NN and neuron.  From a mathematical point of view, neurons are made of compositions of a nonlinear function (also called activation function) and a linear function \cite{bGS}. Therefore, NNs which are a network of neurons can be considered as a nonlinear function. This survey focuses on the results on feedforward neural networks (FNNs) aka multi-layer perceptrons (MLPs) which are the simplest and most popular category of NNs.    
 Nowadays, NN is one of the most trending areas in \textit{artificial intelligence} (AI) and \textit{machine learning} (ML) because of its ability to model complicated relationships \cite{bSS,bIG,bMZ}. UATs are theorems associated with the approximation capabilities of NNs i.e., the ability of an NN to approximate arbitrary functions \cite{bDTL}.  In general, UATs imply that NNs with appropriate parameters can approximate any continuous functions, i.e. are generalized models that can represent complicated relationships in the data \cite{bKF,bKH,bGC}. Further, UATs also focus on the expressivity of NNs which describes the class of functions an NN (with a given structure) can approximate \cite{bMR,bZL}. The structure of an NN can be characterized by the number of layers in an NN aka \textit{depth}, and the maximum number of neurons in a layer aka \textit{width}.  There exist a few surveys on UATs \cite{bDTL,bAK1,bAK2} which focus on specific areas such as soft computing, differential geometry, etc. The initial surveys on UATs \cite{bDTL,bAP99} cover the results from arbitrary width direction only.
Even though the approximation capability of NNs is of both theoretical and practical interest, a comprehensive survey that covers results from arbitrary width and depth directions is absent in the literature to the best of the author's knowledge.  This motivates this paper in which the main objective is to provide a detailed survey on UATs which is useful to students and researchers interested in this topic.
\par The paper is organized as follows. Section 2
gives a brief introduction to NNs and the terminologies associated with NNs. Section 3 starts with the preliminary results on function approximation which is followed by UATs in the direction of arbitrary width and arbitrary depth. 
Finally, section 4 gives the conclusions and further reading.

\par \textit{Notations and definitions:}
$\mathbb{N}$ and $\mathbb{R}$ denote the set of natural numbers and real numbers, respectively.
$\mathbb{R}^{n}$ represents $n$ - dimensional Euclidean space and $\mathbb{R}^{m \times n}$ refers to the space of $m \times n$ real matrices. 
Matrices and vectors are denoted by boldface letters ($\textbf{A},\textbf{a}$),  scalars by normal font ($A,a$), and sets by blackboard bold font ($\mathbb{A}$). The $p-$ norm or $l^{p}$ norm of a vector $\textbf{x} \in \mathbb{X} \subseteq \mathbb{R}^{n}$ is defined as:
\begin{equation}
\label{eqnorm} 
    \parallel \textbf{x} \parallel_{p} = \big( |x_{1}|^{p} + |x_{2}|^{p}+...+|x_{n}|^{n} \big)^{\frac{1}{p}}.
\end{equation}
The $L^{p}$ norm of a function $\textbf{f}:\mathbb{X}\rightarrow \mathbb{Y}$ is defined as
\begin{equation}
\label{eqfpnorm} 
  \parallel \textbf{f} \parallel_{p} =  \bigg(\int_{\mathbb{X}} {\parallel \textbf{f}(\textbf{x}) \parallel}_{p}^{p} d\textbf{x} \bigg)^{\frac{1}{p}}.
\end{equation}
Lebesgue spaces or $\mathbb{L}^{p}$ spaces are function spaces containing measurable functions $\textbf{f}$ satisfying:
\begin{equation}
\label{eqLp} 
    \parallel \textbf{f} \parallel_{p} < \infty.
\end{equation}
A subset $\mathbb{X}$ of $\mathbb{R}^{n}$ is said to be compact if it is closed and bounded.
A subset $\mathbb{S}$ of a topological space $\mathbb{T}$ is said to be \textit{dense} in $\mathbb{T},$ if every element of $\mathbb{T}$ either belongs to $\mathbb{T}$ or arbitrarily close to an element of $\mathbb{T}.$ Examples of Topological spaces are Euclidean spaces, metric spaces, manifolds, etc.
 
 \section{Neural Network (NN)}
\begin{figure}[H]
 		\begin{center}
 		\includegraphics [scale=.35] {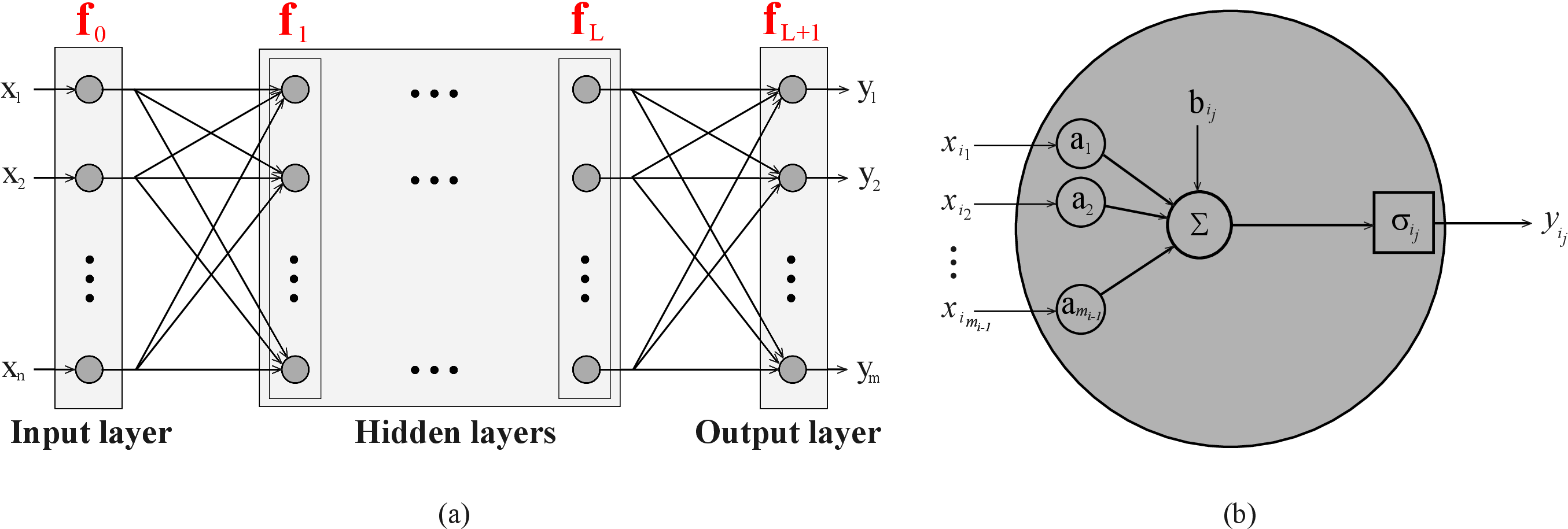}
 		\caption{{ (a) Neural Network \hspace{0.2cm} (b) Neuron.}}
   \label{figNN}
 	\end{center}
 \end{figure}

 NN is a network of neurons arranged in layers.  The layers are connected sequentially, i.e.,  the output of each layer goes to the next layer as input. This results in a network or multi-layer configuration.
Consequently, NNs can  represented using  $\textbf{composite function}$s of the form \cite{bMA1}:
\begin{equation}
\label{eqNN}
\textbf{y}= \textbf{f}_{\text{NN}}(\textbf{x})=\textbf{f}_{\text{L+1}}(\textbf{f}_{\text{L}}...\textbf{f}_{1}(\textbf{f}_{0}(\textbf{x}))) 
\end{equation}
where $\textbf{x}\in \mathbb{R}^{n}$ is the input, $\textbf{y}\in \mathbb{R}^{m}$ is the predicted output, $\textbf{f}_{\text{NN}}:\mathbb{R}^{n} \rightarrow \mathbb{R}^{m} $ is the NN function, and $\textbf{f}_{0},\textbf{f}_{1},...,\textbf{f}_{\text{L+1}}$ are $\textbf{layers}$ of the NN. Here $\textbf{f}_{0}$ represents the input layer, $\textbf{f}_{\text{L+1}}$ is the output layer, and $\textbf{f}_{1},...,\textbf{f}_{\text{L}}$ are the hidden layers (see Fig. \ref{figNN}(a)). 

The number of hidden layers is denoted by $\text{L}$ and the total number of layers in the NN is $\text{L+2},$ i.e., including the input and output layer. Let the input to $i^{th}$ layer is denoted as $\textbf{x}_{i}$ and the output of  $i^{th}$ layer as $\textbf{y}_{i}.$ 
Then each layer of the NN can be represented as: 
\begin{equation}
\label{eqlayer} 
    \textbf{y}_{i}=\textbf{f}_{i}(\textbf{x}_{i})=\bm{\sigma}_{i}(\textbf{A}_{i}\textbf{x}_{i}+\textbf{b}_{i})
\end{equation}
where $\bm{\sigma}_{i}=\left[\begin{matrix}   \sigma_{i_1} \\    \vdots \\   \sigma_{i_{m_i}} \end{matrix}\right]$ contains the element-wise activation functions for the $i^{th}$ layer,  $m_{i}$ is the number of neurons in $i^{th}$ layer, $\textbf{A}_{i}\in \mathbb{R}^{m_{i} \times m_{i-1}}$ is the weight matrix, $\textbf{b}_{i} \in \mathbb{R}^{m_i}$ is the bias vector,  $\textbf{x}_{i}\in \mathbb{R}^{m_{i-1}}$ and $\textbf{y}_{i} \in \mathbb{R}^{m_i}$. Note that the layers in Eq. (\ref{eqlayer}) itself are composite functions and the NN is a composition of layers which also becomes a composite function. The affine term $\textbf{A}_{i}\textbf{x}_{i}+\textbf{b}_{i}$ is the linear component of NNs and the activation function is the nonlinear component.  
In general, any real-valued function can be used as an activation function.  The following are the commonly used activation functions \cite{bAA}:
 \begin{enumerate}
 
 \item $\textbf{ReLU}$ (Rectified Linear Unit) function:
    \begin{equation}
        \sigma(x)=\max(0,x)= \begin{cases}
        0, \hspace{0.3cm} & if \hspace{0.1cm} x\leq 0\\
        x, \hspace{0.3cm} &if \hspace{0.1cm} x >0.
        \end{cases}
    \end{equation}

 \item $\textbf{Step}$ function:
     \begin{equation}
        \sigma(x)=\begin{cases}
        0, \hspace{0.3cm} & if \hspace{0.1cm} x\leq 0\\
        1, \hspace{0.3cm} &if \hspace{0.1cm} x >0.
        \end{cases}
     \end{equation}  

 \item $\textbf{Logistic}$ function:
    \begin{equation}
      \sigma(x)=\frac{1}{1+e^{-x}}.
    \end{equation}

 \item $\textbf{Tanh}$ (Hyperbolic tangent) function:
    \begin{equation}
      \sigma(x)=tanh(x)=\frac{e^{x}-e^{-x}}{e^{x}+e^{-x}}.
    \end{equation}

 \end{enumerate}

\begin{figure}[H]
 		\begin{center}
 		\includegraphics [scale=.4875] {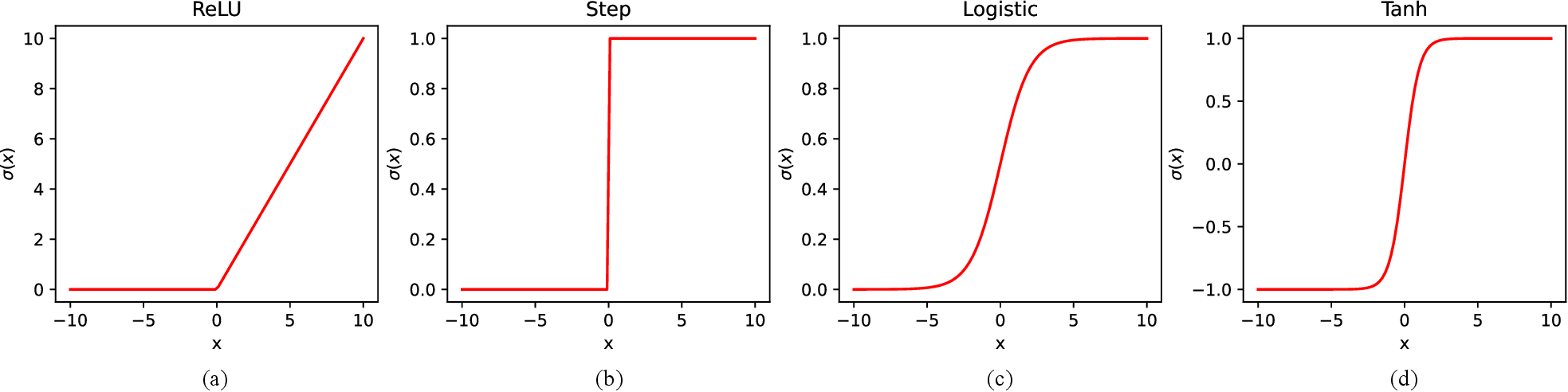}
 		\caption{{ Graph of activation functions: \hspace{.1cm} (a) ReLU \hspace{.1cm}(b) Step \hspace{.1cm}(c) Logistic \hspace{.1cm}(d) Tanh.}}
   \label{figAF}
 	\end{center}
 \end{figure}

The graph of these functions is given in Fig. \ref{figAF}. Note that logistic and tanh functions come under the category of \textit{sigmoid} functions. In general, a sigmoid function is a bounded function (normal range [0,1]) that has a non-negative derivative at each point in $\mathbb{R}$ and exactly one inflection point, i.e., sigmoid functions are characterized by an $S-$shaped curve as in Figs. \ref{figAF}(c) and (d).

 Terminologies associated with NNs and their brief description are given below:
\begin{itemize}
    \item $\textbf{Neuron}$: is the data processing units in NNs.
Neurons can be represented as a scalar-valued function:
    \begin{equation}
    y_{i_j}=\sigma_{i_j}(\textbf{A}_{i_j}\textbf{x}_{i}+b_{i_j})
\end{equation}
where $y_{i_j}$ is the output of $j^{th}$ neuron in the $i^{th}$ layer, $\textbf{A}_{i_j}=\left[\begin{matrix}   a_{1} & a_{2}  &  \dots  a_{m_{i-1}} \end{matrix}\right]$ is the $j^{th}$ row of the weight matrix $\textbf{A}_{i},$ and $b_{i_j}$ is the $j^{th}$ element of the bias vector $\textbf{b}_{i}$ (see Fig. \ref{figNN}(b)). 

\item ${\textbf{Layer}}$: is a collection of neurons that takes the same inputs. In general, layers are vector-valued functions.  Layers can grouped into three categories: input, output, and hidden layers.

\item   $\textbf{Input layer}$: in which each neuron has only one input and it simply passes the input, i.e., $\bm{\sigma}_{0}(\textbf{A}_{0}\textbf{x}+\textbf{b}_{0})=\textbf{x}.$
   \item  $\textbf{Hidden layer}$: in which each neuron can have more than one input and gives a scalar output. The activation function is normally nonlinear.
   \item   $\textbf{Output layer}$: in which each neuron can have more than one input and gives a scalar output. The activation function is normally linear for regression problems and nonlinear in the case of classification problems.

   \item   $\textbf{Width}$:  is the maximum number of neurons in a layer, i.e., width $\text{W} = \underset{i}{\max}\hspace{.2cm} m_{i}.$ 
   
 \item   $\textbf{Depth}$:  is the number of hidden layers plus output layer, i.e., depth $\text{D =  L+1}$. 

 \item   $\textbf{Shallow NN}$: is an NN with only one hidden layer, i.e., $\text{D}=2.$

\item   $\textbf{Deep NN}$: is an NN with more than one hidden layer, i.e., $\text{D}>2.$
   
 \end{itemize} 

\begin{figure}[H]
 		\begin{center}
 		\includegraphics [scale=.4875] {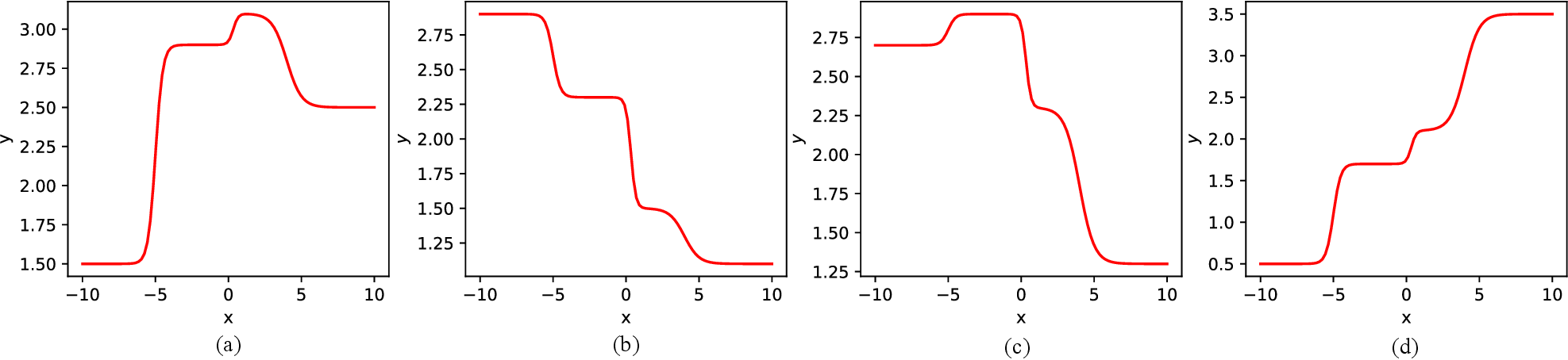}
 		\caption{{ Illustrating NN.}}
   \label{figNNeg}
 	\end{center}
 \end{figure}

 As an example, we can consider an NN with one input, one output, a linear output layer, and one hidden layer with three neurons which can be represented as 
 \begin{equation}
 \label{eqy2}
     y=\textbf{A}_{2} \bm{\sigma}_{1}(\textbf{A}_{1}x+\textbf{b}_{1})+b_{2}.
 \end{equation}
Let $\textbf{A}_{2}=\left[\begin{matrix}   0.1 &   0.3 &  0.7 \end{matrix}\right], \textbf{A}_{1}=\left[\begin{matrix}   3 \\   -1 \\  2 \end{matrix}\right],$ $\textbf{b}_{1}=\left[\begin{matrix}   -1 \\   4 \\  10 \end{matrix}\right],$ $b_{2}=2,$ $\bm{\sigma}_{1}=\textbf{tanh}$ which will expand Eq. \ref{eqy2} as:
\begin{equation}
    \label{eqy2exp}
    y=0.1 \hspace{.1cm} tanh (3x-1) + 0.3 \hspace{.1cm} tanh (-x+4) + 0.7 \hspace{.1cm} tanh (2x+10)+2
\end{equation}
and for $x \in [-10,10]$ we can compute the output by substituting in Eq. (\ref{eqy2}). Fig. \ref{figNNeg}(a) shows the plot of the output $y$. 

Now by changing $\textbf{A}_{2}=\left[\begin{matrix}   -0.4 &   0.2 &  -0.3 \end{matrix}\right]$ and keeping $\textbf{A}_{1},b_{1},b_{2}$ as before, we get the output as in Fig. \ref{figNNeg}(b). Similarly for $\textbf{A}_{2}=\left[\begin{matrix}   -0.3 &   0.5 &  0.1 \end{matrix}\right]$ and $\textbf{A}_{2}=\left[\begin{matrix}   0.2 &   -0.7 &  0.6 \end{matrix}\right]$ the output will be as in Figs. \ref{figNNeg}(c) and (d), respectively. These examples illustrate that we can represent complicated nonlinear relationships using NNs. As the main use of NNs is modeling relationships in the data, the question of the approximation capabilities of NNs gained interest naturally. Now the question is \textit{can we approximate any continuous function using NNs?} This leads to the theoretical results known as UATs which are discussed next.   
\section{Universal Approximation Theorems (UATs)}

\begin{figure}[H]
 		\begin{center}
 		\includegraphics [scale=.35] {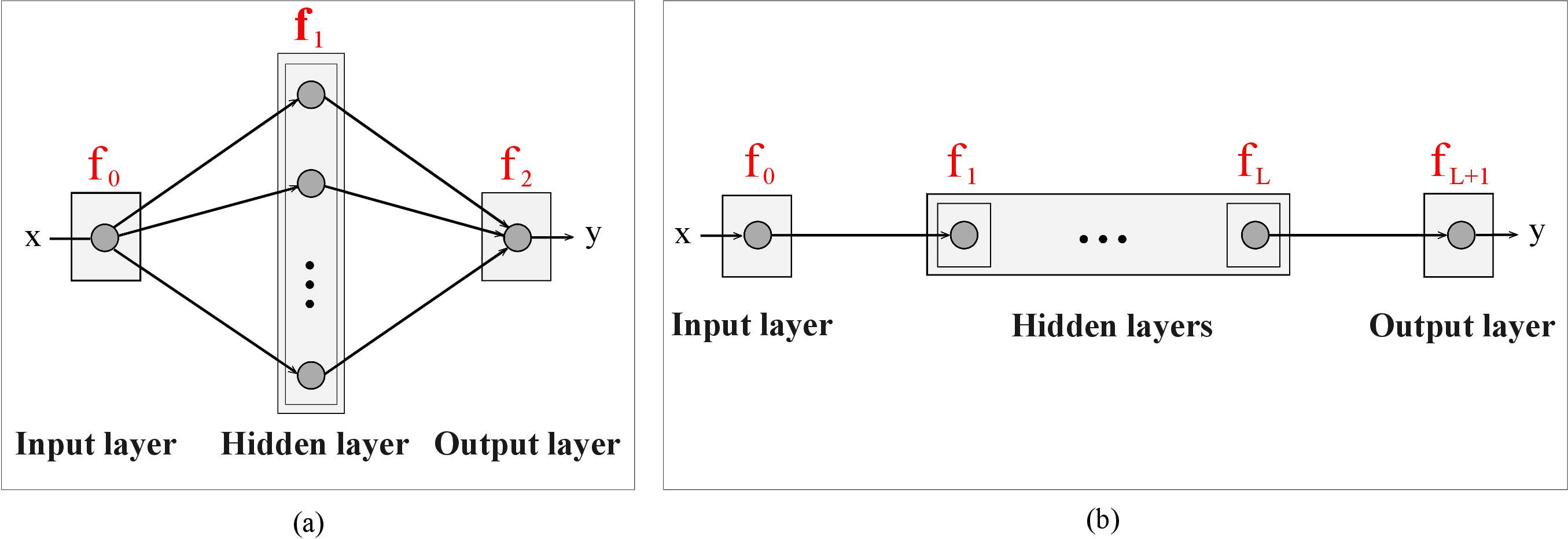}
 		\caption{{ (a) NN with arbitrary width (b) NN with arbitrary depth.}}
   \label{figNNA}
 	\end{center}
 \end{figure}

Universal approximation theorems (UATs) are theoretical results associated with the approximation capabilities of NNs. In general, UATs are explored in the following directions: 
\begin{enumerate}
    \item \textbf{Arbitrary width case}: deals with approximation capabilities of NNs with an arbitrary number of neurons (with a limited number of hidden layers). For eg. Fig \ref{figNNA}(a) shows an NN with one hidden layer that has arbitrary neurons.
    \item \textbf{Arbitrary depth case}: is associated with NNs with an arbitrary number of hidden layers (with a limited number of neurons). For eg.  Fig \ref{figNNA}(b) shows NN with arbitrary hidden layers and one neuron per layer.
\end{enumerate}
Before moving on to UATs in arbitrary width and depth cases, a brief summary of the preliminary results on function approximation is given next.

\subsection{UATs: Predecessors}
 
 The ability of NNs to approximate arbitrary functions with a desired accuracy has theoretical and practical significance. The theoretical interests originated from the mathematical areas: linear algebra \cite{bGS}, analysis \cite{bAB},   topology \cite{bGB}, etc. The problem of representing a continuous function as a superposition of simpler components (also known as basis functions) has been studied long back.
One of the initial results in this direction is the
Taylor's theorem\cite{bBT,bEM} which is stated below:
\begin{theorem}[Taylor, 1715]
\textit{ Any continuous function $f(x):\mathbb{R} \rightarrow \mathbb{R}$ that is $k-$ times differentiable at $a\in \mathbb{R}$ can be represented as a sum of polynomials:
    \begin{equation}
    \label{eqTAY} 
        f(x)=f(a)+f^{'}(a)(x-a) +\frac{f^{''}(a)}{2}(x-a)^{2}+...+\frac{f^{k}(a)}{k!}(x-a)^{k}+R_{k}(x)=\sum_{i=0}^{k} c_{i} (x-a)^{i}+R_{k}(x)
    \end{equation}
  where $c_{i}=\frac{f^{i}(a)}{i!}= \frac{1}{i!} \frac{d^{i}}{dx^i}f(x)|_{x=a} $ and  $R_{k}(x)=o(|x-a|^{k})$ is the residual term.}   
\end{theorem}
A special case of the Eq. \ref{eqTAY} for $a=0$ gives:
  \begin{equation}
        f(x)=\sum_{i=0}^{k} c_{i} x^{i}+R_{k}(x)
   \end{equation}
which is a more commonly used version and the summation is called the Maclaurin series. Taylor/Maclaurin series shows that polynomial functions can approximate any continuous smooth functions with sufficient accuracy.
  Taylor's theorem can be considered as a mathematical foundation for linearization-based methods. 
Similar ideas were presented in the
Fourier series\cite{bJF,bAP} in which the results are initially given for periodic functions:
\begin{theorem}[Fourier, 1807]
  \textit{  Any continuous and periodic function $f(x)$ can be expressed as a sum of sinusoids:}
    \begin{equation}
    \label{eqFR} 
        f(x)=A_{0}+\sum_{i=1}^{N}A_{i}cos(\frac{2\pi i}{T} x + \phi_{i}).
    \end{equation}
    \textit{ where $T$ is the period of the function, $A_{i}$ is the amplitude and $\phi_{i}$ is the phase of $i^{th}$ harmonic component. } 
\end{theorem}
Later on, the results are generalized for nonperiodic functions as well, which resulted in the  Fourier transform.
The Fourier series and transform resulted in the origin of frequency domain analysis and design methods. Another milestone in this area is the Weierstrass approximation theorem \cite{bKW} which can be considered as an extension of Taylor's theorem to arbitrary continuous functions and is stated below:
\begin{theorem}[Weierstrass, 1885]
  \textit{  Any continuous real-valued function $f(x):[a,b] \rightarrow \mathbb{R}$ defined on the interval $[a, b]$ can be approximated  with a polynomial function $p_{N}(x)=\sum_{i=0}^{N} c_{i} x^{i}$ with finite degree $N$ such that: 
    \begin{equation}
    \label{eqWR} 
        |f(x)-p_{N}(x)| < \epsilon
    \end{equation}
    for arbitrary $\epsilon>0$}. 
\end{theorem}
Weierstrass's theorem implies that any continuous function on a closed interval can be uniformly approximated by a polynomial function with arbitrary accuracy. The Weierstrass approximation theorem can be considered as a mathematical foundation for polynomial regression and interpolation methods. 

Later on in the 1900s, the problem of decomposing continuous multivariable functions as a finite superposition of
 continuous univariate and bivariate functions attracted a lot of research interest. This started with the continuous variant of Hilbert's $13^{th}$ problem \cite{bDH}: \textit{Can any continuous function of more than two variables be expressed as a superposition of finitely many continuous functions of two variables?} This problem was solved by Arnold and Kolmogorov in 1957 which resulted in the Kolmogorov-Arnold representation theorem \cite{bAK}  stated below:

 \begin{theorem}[Kolmogorov and Arnold, 1959]
  \textit{ Any continuous multivariate function $f: [0,1]^{n}\rightarrow \mathbb{R}$  
   can be written as}
   \begin{equation}
       \label{eqKA} f(\textbf{x})=f(x_{1},x_{2},...,x_{n})=\sum_{j=1}^{2n+1} \beta_{j}\bigg(\sum_{i=1}^{n}\alpha_{ij}(x_{i})\bigg)
   \end{equation}
  \textit{ where $\alpha_{ij}:[0,1]\rightarrow \mathbb{R}$ and $\beta_{j}:\mathbb{R} \rightarrow \mathbb{R}.$}
 \end{theorem}

Theorems 1 to 4 above gave insights into the approximation capabilities of sinusoidal functions, polynomial functions, etc. 
After the introduction of NNs, approximation capabilities of sigmoid functions gained popularity, since in the initial versions of NNs sigmoid functions were used as activation functions. The approximation capabilities of NNs with different activation functions such as ReLU, step, and tanh were introduced later. This resulted in several theorems known by the name of UATs which are discussed in the rest of the paper.

\subsection{UATs: Arbitrary width case}
\par
\begin{figure}[H]
 		\begin{center}
 		\includegraphics [scale=.4875] {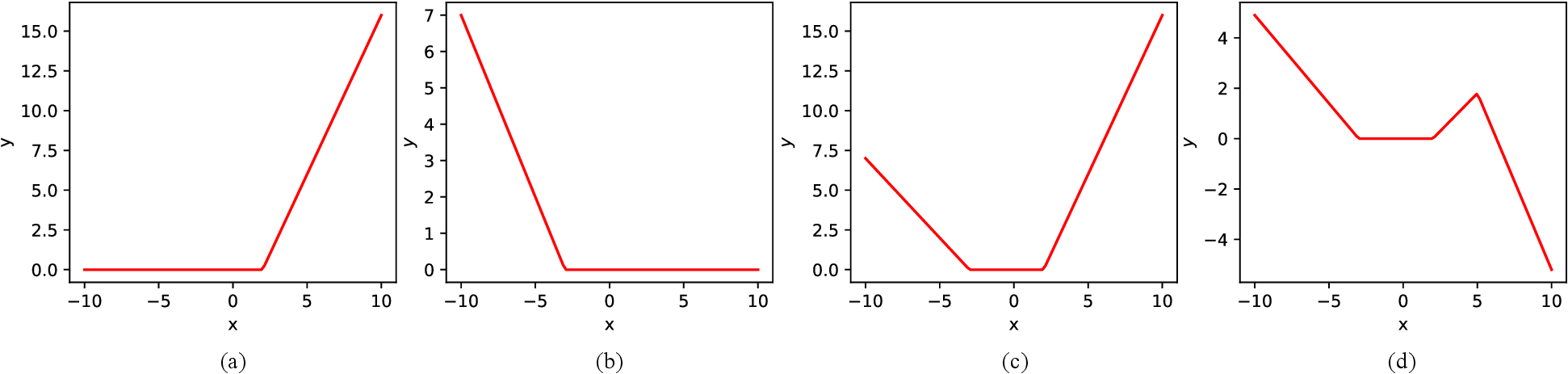}
 		\caption{{ Output of NNs with one hidden layer and ReLU activation function.}}
   \label{figNNO}
 	\end{center}
 \end{figure}

 This section discusses results on approximation capabilities of NNs with arbitrarily large widths.  The depth or number of layers in the NN is considered to be bounded.  
As an example, consider the following NNs with one hidden layer with ReLU activation functions and a linear output layer:
\begin{equation}
\label{eqAW} 
    \begin{aligned}
      &(a) \hspace{.5cm} y=\text{ReLU}(2 x - 4)\\
        &(b) \hspace{.5cm}y=\text{ReLU}(- x- 3)\\
        &(c) \hspace{.5cm}y=\text{ReLU}(2 x - 4)+\text{ReLU}(- x - 3)\\
        &(d) \hspace{.5cm}y=0.3\hspace{0.05cm}\text{ReLU}(2 x -4)+0.7\hspace{0.05cm}\text{ReLU}(- x- 3)-0.5\hspace{0.05cm}\text{ReLU}(4 x - 20)
    \end{aligned}
\end{equation}
which corresponds to NNs with one hidden layer neuron (a,b), two neurons (c), and three neurons (d). For these cases,
 the outputs are shown in Fig. \ref{figNNO}. The plots show that the number of folds in the curve increases with the number of neurons. For example, the curve in Fig. \ref{figNNO} (c) and (d) has two and three folds which correspond to NNs with two and three hidden layer neurons, respectively.  This indicates that an arbitrary number of folds can be introduced to the curve by an NN with arbitrary neurons. 
 Folded lines are also called a piecewise linear curve. Therefore, with one hidden layer and ReLU activation function, we can obtain piecewise linear approximations of continuous functions.
A theoretical result on the number of folds and linear pieces is presented in \cite{bPMB13} which is stated below:

\begin{theorem}[Pascanu et al., 2013]
    Consider the graph of $\textbf{f}_{\text{NN}}(\textbf{x})$   which has  folds along $H$ hyperplanes
defined linear equations $\textbf{A}_{1_j}\textbf{x} + b_{1_j} = 0,$ $\textbf{x}\in \mathbb{R}^{n},$ $j=1,2,...,H$. Then the number of linear pieces of $\textbf{f}_{\text{NN}}$ is:
\begin{equation}
    n_{l} =  \sum_{i=0}^{n} {^H}{C_i}.
\end{equation}
\end{theorem}

In general, the approximation capabilities (the number of folds or bends in the outputs increases) as the number of neurons in the hidden layer increases which can observed from Fig. \ref{figNNO}. This makes NNs with a sufficient number of neurons and hidden layers a universal approximator that can used for modeling data with arbitrary patterns, i.e., NNs are generalized models. This is the central idea of UATs. 
\par The initial versions of UATs were introduced during the 1980s which attempted to extend the Kolmogorov-Arnold representation theorem and other preliminary results to NNs.
In Cun \cite{bYC87} and Lapedes and 
Farber \cite{bLF88} it has been shown that continuous functions can be arbitrarily approximated using NNs with \textit{two hidden layers} and monotonic activation functions. 
Irie and Miyake \cite{bIM88} showed that arbitrary 
functions can be approximated by a one hidden layer network with an \textit{infinite number of neurons}.
Later on, Gallant 
and White \cite{bGW88} showed one hidden layer with 
a monotone cosine activation function can give Fourier series approximation to a given function as its output. However, this approximation is limited to only square-integrable functions
on a compact set (see Dirichlet's conditions \cite{bPD29} for Fourier series approximation). Later on generalizations of these results to arbitrary continuous functions in a compact set were introduced by Funahashi \cite{bKF}, Hornick et al. \cite{bKH}, and  Cybenko \cite{bGC} which are the popular versions of UATs in the arbitrary width case. The  UATs are mostly stated using the density of the functions $\textbf{f}_{\text{NN}}(\textbf{x})$ generated by NNs within a given function space of interest, i.e. if $\textbf{f}_{\text{NN}}(\textbf{x})$ is dense in a given space, then, any function in that space can be approximated by $\textbf{f}_{\text{NN}}$. This leads to the following theorem:
\begin{theorem}[Funahashi,  Hornick et al., and Cybenko,  1989]
\textit{ Let $\mathbb{X}$ be any compact subset of $\mathbb{R}^{n}$ and $\sigma$ be any sigmoid activation function, then the finite sum of the form}: 
 \begin{equation}
\label{eqCY} 
 f_{\text{NN}}(\textbf{x}) = \textbf{A}_{2} \bm{\sigma}(\textbf{A}_{1}\textbf{x}+\textbf{b}_{1})=  \sum_{j=1}^{m_{1}} a_{2_j} \sigma(\textbf{A}_{1_j}\textbf{x}+b_{1_j})
 \end{equation}
\textit{ is dense in $\mathbb{X}.$ 
In other words, given any $f:\mathbb{X}\rightarrow \mathbb{R}$ and $\epsilon > 0$, there is a finite sum:} $f_{\text{NN}}$ \textit{as in Eq. (\ref{eqCY}) for which}  $|f(\textbf{x})-f_{\text{NN}}(\textbf{x})| < \epsilon$ 
\textit{for all} $\textbf{x} \in \mathbb{X}$. 
\end{theorem}
The above theorem means that NNs with one
hidden layer and sigmoid activation function can approximate any continuous univariate function on a bounded domain
with arbitrary accuracy.  Hornik \cite{bKH2} showed that the primary reason behind the universal approximation capabilities of NNs is
  the multilayer feed-forward architecture. 
 In \cite{bML} it is shown that MLP with non-polynomial activation functions are universal approximators which resulted in the following theorem:
\begin{theorem}[Leshno et al., 1993]
    Let $\mathbb{X}$ be any compact subset of $\mathbb{R}^{n}$ and $\sigma$ be an activation function, then the finite sum in Eq. \ref{eqCY} is dense in $\mathbb{X}$ iff $\sigma$ is not a polynomial function. 
\end{theorem}
Later on, similar theorems were developed for other activation functions such as ReLU, step, tanh, etc which are non-polynomial functions.
Therefore, depth-2 NNs (NNs with one hidden layer aka shallow NNs) with a suitable activation function are universal approximators. The same holds for NNs with more than one hidden layer aka deep NNs.
However, depth-1 NNs or NNs without any hidden layer have limited approximation capability. Further results of UATs for shallow NNs can be found in \cite{bNG,bJP}.

\subsection{UATs: Arbitrary depth case}

\begin{figure}[H]
 		\begin{center}
 		\includegraphics [scale=.475] {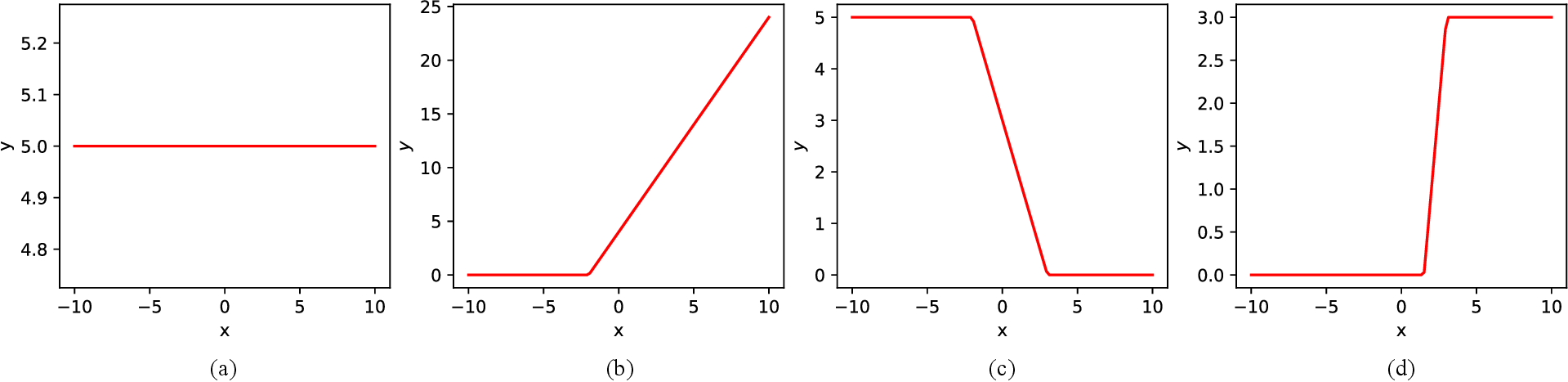}
 		\caption{{Output of NNs with one neuron per layer.}}
    \label{figDNN}
 	\end{center}
 \end{figure}

This section focuses on the approximation capabilities of deep NNs, i.e. NNs with arbitrary depth and bounded width. The arbitrary depth case has attained a lot of research interest recently, especially after the introduction of deep learning as a separate domain in ML. 
To get a better understanding of this idea, let's start with NNs with one neuron per layer, i.e. NNs with width $\text{W}=1$ and arbitrary depth as shown in Fig. \ref{figNNA}(b). In that case, each layer will be a scalar-valued function. Consider the following NNs with ReLU activation functions in the hidden layer and a linear output layer:
\begin{equation}
\label{eqAW} 
    \begin{aligned}
      &(a) \hspace{.5cm} y=\text{ReLU}(0 x+ 5)\\
        &(b) \hspace{.5cm}y=\text{ReLU}(2 x+ 4)\\
        &(c) \hspace{.5cm}y=\text{ReLU}\big(-0.5 [\text{ReLU}(2 x+ 4) ]+ 5 \big)\\
        &(d) \hspace{.5cm}y=\text{ReLU}\big( -2 \big[\text{ReLU}\big(-0.5 [\text{ReLU}(2 x+ 4) ]+ 5 \big) \big] + 3 \big)
    \end{aligned}
\end{equation}
which corresponds to NNs with one hidden layer (a,b), two hidden layers (c), and three hidden layers (d). For these cases,
 the outputs are shown in Fig. \ref{figDNN}.
 
 For NNs with one neuron per layer and ReLU activation function, the output can be either a straight line as in Fig. \ref{figDNN}(a), piecewise linear curve with one fold as in Fig. \ref{figDNN}(b), or piecewise linear curve with two folds as in Fig. \ref{figDNN}(c) and (d). 
The tilted portion in Fig \ref{figDNN}(c) can be rotated by multiplying it with a weight, which changes the slope. However,
    the horizontal portions in Fig. \ref{figDNN}(c) correspond to a constant output. Therefore, multiplying it with a weight term gives another constant output. This means the horizontal portions in Fig. \ref{figDNN}(c) can not be rotated with a linear mapping.  Consequently, after two folds, the subsequent layers cannot add any further folds which can be observed in \ref{figDNN}(d). However, by increasing the number of neurons in the hidden layer as in Eq. (\ref{eqAW}), the number of folds can be increased beyond 2. This shows that width-1 NNs with ReLU activation have limited approximation capabilities. This holds for other activation functions as well. A theoretical result on this aspect is presented in \cite{bZL} which is stated below:
    
\begin{theorem}[Lu et al., 2017]
   \textit{  Except for a negligible set, all functions $f: \mathbb{R}^{n}\rightarrow \mathbb{R}$ cannot be
 approximated by any ReLU network whose width $W\leq n$.}
\end{theorem}

The above theorem states that width-1 NNs can approximate only a small class of univariate functions, i.e., the minimum width required for universal approximation should be greater than 1. Similar results can be found in \cite{bJJ}. This leads to the problem of finding the \textit{minimum width} for universal approximation with deep NNs. One of the initial results in this direction states that $width-n+4$ NNs with ReLU activation function are universal approximators which is stated below \cite{bZL}:  
\begin{theorem}[Lu et al., 2017]
\textit{ For any
 Lebesgue-integrable function $f : \mathbb{R}^{n} \rightarrow \mathbb{R}$ and $\epsilon>0,$  there exists a neural network} $f_{\text{NN}}$ \textit{of width $W\leq n+4$ with ReLU activation function which satisfies: }
\begin{equation}
 \label{eqLU} 
   \int |f(\textbf{x})-f_{\text{NN}}(\textbf{x})|d\textbf{x} < \epsilon.
\end{equation} 
 
\end{theorem}
This means that NNs with arbitrary hidden layers and at most $n+4$ number of neurons per layer can approximate any functions in a Lebesgue integrable space with sufficient accuracy.   Later on, it is shown that the minimum width required for universal approximation in deep NNs is $n+1$ which leads to the following theorem \cite{bSP}: 
\begin{theorem}[Park et al., 2021]
\textit{ The minimum width required for universal approximation of Lebesgue integrable
functions  $\textbf{f}:\mathbb{R}^{n} \rightarrow \mathbb{R}^{m}$ is  $max \{n+1,m\}.$}
\end{theorem}
Therefore, width-2 NNs with a suitable activation function are also universal approximators for continuous univariate functions. Similar results of UATs for width-bounded NNs can be found in \cite{bGG,bTP}. Apart from this, UATs for bounded-depth and bounded-width NNs are also studied in the literature \cite{bMP99,bHY}.

\section{Conclusions and Further reading}
This paper discussed the major results and theorems on the approximation capabilities of NNs. The paper focused on UATs for feedforward NNs or MLPs. The extension of UATs for other version of NNs such as convolutional NN \cite{bDZ,bAH},  ResNet \cite{bPT,bHL}, recurrent NN \cite{bSH23,bXCY}, transformer \cite{bSAN}, etc are available in the literature.
For other topics on NN and ML, the books \cite{bGS,bMZ} can referred to. The Python codes for the examples discussed in this tutorial are available at GitHub \footnote{\url{https://github.com/MIDHUNTA30/NN-PYTHON}}.

\bibliographystyle{unsrt}

\end{document}